# Robust Facial Expression Classification Using Shape and Appearance Features

S L Happy and Aurobinda Routray
Department of Electrical Engineering,
Indian Institute of Technology Kharagpur, India

*Abstract*— **Facial expression recognition has many potential applications which has attracted the attention of researchers in the last decade. Feature extraction is one important step in expression analysis which contributes toward fast and accurate expression recognition. This paper represents an approach of combining the shape and appearance features to form a hybrid feature vector. We have extracted Pyramid of Histogram of Gradients (PHOG) as shape descriptors and Local Binary Patterns (LBP) as appearance features. The proposed framework involves a novel approach of extracting hybrid features from active facial patches. The active facial patches are located on the face regions which undergo a major change during different expressions. After detection of facial landmarks, the active patches are localized and hybrid features are calculated from these patches. The use of small parts of face instead of the whole face for extracting features reduces the computational cost and prevents the over-fitting of the features for classification. By using linear discriminant analysis, the dimensionality of the feature is reduced which is further classified by using the support vector machine (SVM). The experimental results on two publicly available databases show promising accuracy in recognizing all expression classes.**

*Index Terms*-- **Facial expression recognition, active facial patch, Linear Discriminant Analysis, Local Binary Patterns, Pyramid of Histogram of Gradient, Support Vector Machine.**

## I. INTRODUCTION

Affective computing has become an active field of research due to its wide range of applications in the field of human-machine interaction (HMI), surveillance, health care, personal assistance, law enforcement etc. Works of Ekman [1] provide the correlation of facial expressions with the affective states of a person. Though some researchers use arousal (active/passive) and valence (negative/positive) scale for defining affective state, most of the researchers use the six universal expressions for classifying the expressions, namely: anger, disgust, fear, happiness, sadness and surprise. We propose an expression framework for recognizing six universal expressions by the use of shape and appearance features together from some active regions of face.

Recognizing facial expressions involve the understanding of changes in facial organs such as eyes, cheek, eyebrows, lips etc. Usually, the changes in shape and formation of furrow or wrinkle are prominent observations which can be used for analysis of facial expressions. In Facial Action Coding System (FACS) [2], each muscle movement is considered as an Action Unit (AU), and the combination of different AUs define the facial expression. However, the detection of AUs is a big challenge due to the variation of face shape and texture for different persons. Therefore, researchers depend upon face geometry and appearance parameters [3], [4] for robust recognition of expressions. Geometrical features include shape of the facial organs as well as their relative movements. On the other hand, appearance features represent the texture of local region which changes with different expressions. During an expression, movement of facial organs is always associated with change in appearance of the corresponding region by producing wrinkles, skin folding etc. Therefore, shape features along with the appearance features can construct effective feature vector for robust expression classification.

Among the appearance features, Gabor descriptor [5], [6], and Local Binary Patterns (LBP) [7], [8], [9] are very popular for representing facial features. Recently, several extensions of LBP has been proposed with assuring results for different purposes. However, for computational simplicity and significant accuracy of LBP features has encouraged us to use it in our experiment. Moreover, it is less sensitive to noise and monotonic illumination changes, and performs well with low resolution images.

Automated recognition of facial expression requires minimum intervention of human during processing. However, accurate analysis needs precise face alignment which is yet a challenging task. Moreover, the accurate localization of the facial landmarks depends upon the geometry of face. The standard practice is to divide the face image into different size grids and to extract local features from each sub region as reported in [10] and [11]. This method works well in case of similar size faces and with proper face alignment. However, slight misalignment will cause disturbance in all sub regions, thereby reducing the accuracy. In addition, not all portion of face undergo a prominent change during an expression. Therefore, we have used a few facial patches which are active during different expressions. These active regions lie around the lips, eyes, cheek, eyebrows, and upper nose regions [12]. Zhang et al. [13], reported that the principal components of local facial patches contain enough information to classify the expressions comfortably. In [14], authors used eight facial patches for observing the presence of skin deformations and thereby classifying facial expressions. However, the shape parameters and the appearance parameters were not considered during expression classification.

This paper proposes a robust automated facial expression recognition framework by using shape and appearance features together. Unlike previously adopted methodologies, we introduced facial patch based local shape and appearance feature extraction technique which improves the accuracy of expression recognition. In our approach, the features are extracted from active facial patches based on the position of facial landmark points. Further the features from different sub regions are concatenated to represent the whole image.

The paper is organized as follows. Section II describes the methodology adopted in this paper. Section III explains the results obtained, followed by conclusion and future work in Section IV.

## II. METHODOLOGY

This paper proposes a robust algorithm for facial expression classification by using both shape and appearance features. Based on the training data, the classifiers are constructed by using the combined shape and appearance features and it classifies the expression of an unseen image by extracting the same features out of it.

Face detection is the first step in this approach. Once the face is localized, some facial organs are detected within face region and the active patch locations are determined. We use PHOG features as shape descriptor and histograms of LBP image as appearance features. After extracting geometry and appearance features from each facial patches, they are concatenated to represent the image as a whole. A dimensionality reduction technique is adopted to reduce which is followed by classification of features into different expression categories. The overall idea is presented in Fig. 1.

The proposed algorithm mainly consists of six stages: face detection, detection of facial landmarks, active facial patch localization, feature extraction, dimensionality reduction, and facial expression classification. The rest of the section will describe the algorithms adopted in this paper.

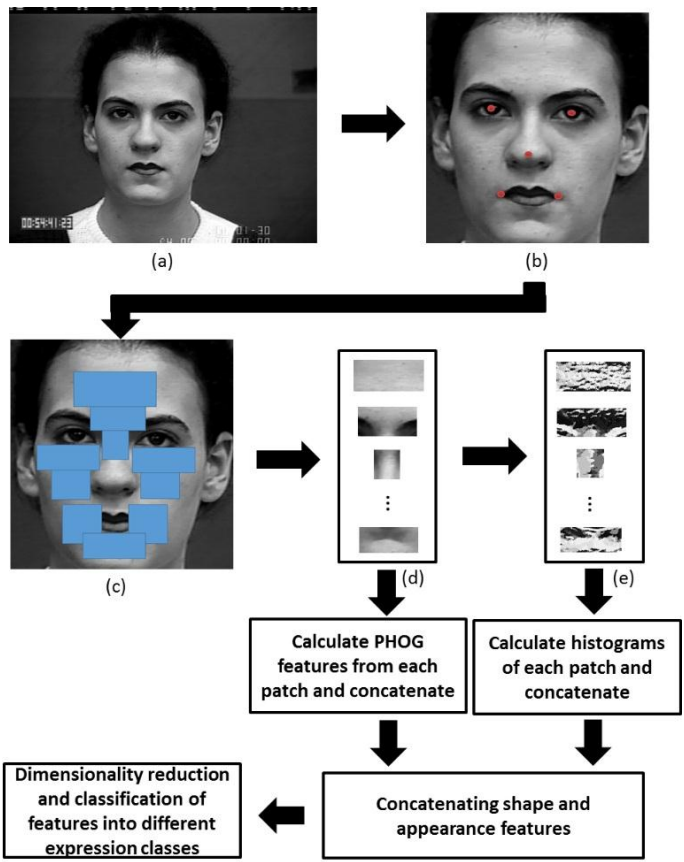

Fig. 1. Conception of the facial expression recognition system, (a) input image, (b) face detection and registration, (c) localization of active facial patches, (d) extraction of facial patches, (e) LBP images of facial patches

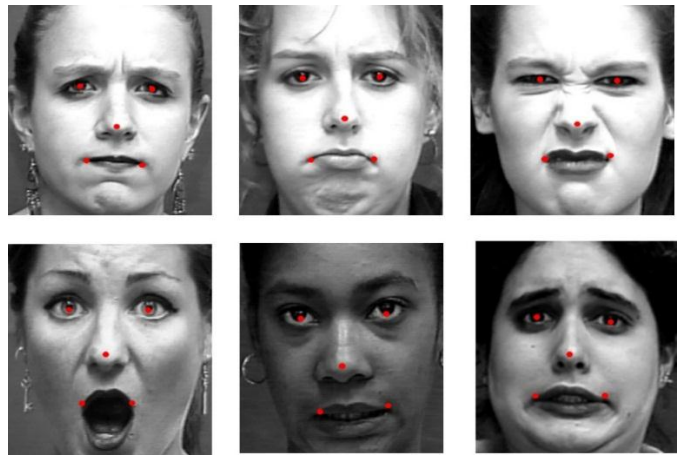

Fig. 2. Some examples of automated landmark detection. Landmarks are properly detected in upper row images, whereas the lips corner is slightly dislocated in lower row images.

### A. Preprocessing

To remove noise during image acquisition, a low pass filtering is carried out first using a Gaussian kernel. Viola-Jones Haar cascade classifiers [15] were used to detect face from the image. Haar classifier based method is chosen for face detection because of its high detection accuracy and real time performance. The face region is extracted and scaled to a common resolution of 96x96.

### B. Detection of Facial Landmarks

The face region is further processed for localization of facial landmark points such as eyes, nose, and lips corners. We have used Haar classifier based detection scheme for localizing eyes and nose. However, the process is optimized by selection of coarse region of interests (ROI) for each organ. For example, the eyes were detected in the upper half of the face image, and nose ROI was at the center of face. The selection of coarse ROIs reduce false alarm rate, while it ensures fast and accurate detection. Similarly, for lips corner detection, we adopted the method reported by Nguyen *et al.* [16]. By searching the high contrast locations from both left and right side of lips ROI, the lips corners were detected. Then the face image is aligned by positioning the eyes horizontally.

### C. Extraction of Active Facial Patches

As reported in [12], the active facial patches usually lie around eyes, lips, eyebrows, and nose. The use of whole face for expression analysis is a drawback since most of the facial regions do not take part in producing facial expressions. Therefore, we selected the face sub regions that undergo a change during different expressions and call it as active facial patches. We have empirically selected 10 active patches in face out of which three were located around the lips, four in cheek, two in forehead and one in upper nose region. The eye regions were not selected as the features from these regions may be redundant for different expressions during person independent analysis. Similarly, inner eyebrow regions were selected instead of the complete eyebrow region. The active facial patch locations are shown in Fig. 1(c). Features around lip corners are very important as it involves in most of the expressions. The patch below the lower lip contribute toward classifying open mouth, tightened lips, and presence of furrows. Similarly, patches in cheek encodes the texture of cheek during smile and disgust. The forehead patches contribute toward classifying the texture of skin around eyebrow region during anger, surprise, fear, sadness etc. The upper nose patch mainly responsible for discriminating disgust and sometimes anger from other expressions.

To locate the active patches, we need to localize some of the facial landmarks such as eyes, nose, and lips corners accurately. Using the geometry of the human face, the patch locations can be derived from these landmark positions using face width as a parameter. Note that the patch positions vary depending upon the face shape for different person. Thus, after localizing the facial landmark points as shown in Fig. 1(b), the active patches are localized which are shown in Fig. 1(c).

### D. Feature Extraction

Accurate expression classification depends upon the proper selection of features. Some of the desired properties of a feature vector are insensitive to scale, rotation, translation, and illumination variation of an image. We have used a combined shape and appearance feature vector for accurate expression classification. As discussed earlier, LBP features are used as appearance features due to its excellent light invariance property and low computational complexity [17]. Similarly, pyramid of histogram of gradients (PHOG) features are used as shape descriptors. After extracting the PHOG and LBP histogram features from each active facial patches, the features are concatenated to represent the whole image.

*1) LBP Histogram Features*

LBP encodes the local texture by using the pixel values of the image. LBP value at a pixel is determined by the neighborhood pixel values by comparing the pixel values of neighbor to the central pixel value. Fig. 3 explains the LBP calculation procedure which is provided in (1).

$$LBP = \sum_{n=0}^{7} s(i_n - i_c) 2^n \quad (1)$$

Where, $i_c$ is the central pixel value and

$$s(x) = \begin{cases} 1, x \geq 0 \\ 0, x < 0. \end{cases}$$

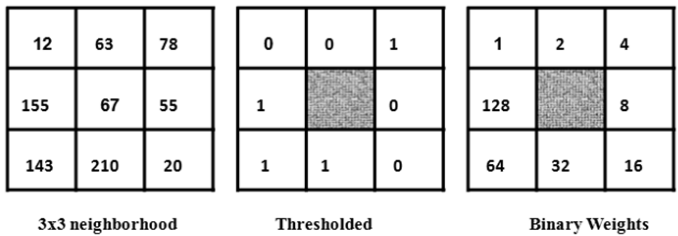

Fig. 3 Calculation of LBP

Further, to compact the representation of the LBP feature, the histograms of LBP image is utilized as feature descriptors, which can be written as

$$H_i = \sum_{x,y} I\{LBP(x,y) = i\}, \quad i = 0,1,\dots,n-1 \quad (2)$$

where $n$ is the number of labels produced by LBP operator. Instead of using a long feature vector, it can be further shortened by using histograms of different binwidth. The LBP features at different patches extracts texture of local region such as presence of wrinkles, furrows, skin deformations etc.

*2) PHOG Descriptor*

To represent shape of an object, edges play important role. However, edges are not good features for classification purpose because it is very sensitive to change of shape, rotation and illumination variation. Therefore, representation of segmented edge directions are widely used for encoding geometry of an object. In histogram of oriented gradients (HOG), the image is divided into several blocks and histograms of edge directions are concatenated as shape descriptor. Here the edge orientation is quantized to different binwidth for angular range 0 ~ 180°. The histogram is calculated by considering the strength of edge at all pixel locations in the sub region.

To represent the position of orientations in addition to the HOG feature, pyramid of histogram of gradients (PHOG) is used which was proposed by Bosch et al. [18]. In PHOG, HOG features of an image is calculated at different grid resolutions and concatenated to achieve more spatial property of the object. Hence, PHOG represents both edge direction and location. Again, PHOG feature is more robust to slight change in object shape unless the overall shape is consistent.

For our purpose, we have used 9 bin histograms for collecting the angular orientations for each 8x8 cell in a patch. A three layer pyramid structure is used for accurately representing the shape.

### E. Dimensionality Reduction

The PHOG and LBP features are concatenated to construct a hybrid feature vector which encodes both shape and texture information. Feature normalization is carried out for proper classification of different images. When the shape and appearance features are concatenated, the dimensionality of the feature vector is increased. Therefore, dimensionality reduction is of prime importance for selection of significant features. Different machine learning algorithms such as principal component analysis (PCA), independent component analysis (ICA), linear discriminant analysis (LDA) etc. can be adopted for dimensionality reduction. For better classification purpose, LDA is used in our experiments.

In LDA, discriminant directions are calculated and the feature vector is projected into the lower dimensional space while protecting the discriminative information. LDA makes sure that the hyper-planes minimize the with-in class scatter while maximizing the between class scatter. Singularity of within scatter matrix creates problem during matrix inversion. To remove possibility of the singularity of the with-in scatter matrix, PCA is applied first to the feature vector as proposed by Belhumeur et al. [19]. Thus an optimal feature space is constructed by using PCA-LDA approach for dimensionality reduction.

## F. Multi-class Expression Classification

Support vector machine (SVM) [20] is used for classifying the expression classes. SVM uses nonlinear mapping to map the feature vector to higher dimensional plane and separates the two classes by a linear decision boundary. However, SVM is a two-class classification technique. So we used one-against-one (OAO) technique for classifying six expression classes. In OAO method, binary classifiers are trained between all pair of classes, i.e. for six classes, $(6*5)/2 = 15$ binary classifiers are trained between all pair of expression classes. Then all the binary classifiers vote to choose the class to which the feature vector belongs to.

## III. EXPERIMENTAL RESULTS

We have used both Japanese Female Facial Expressions (JAFFE) [21] and Cohn-Kanade (CK+) [22] database evaluating the proposed framework. The six universal expression faces are considered for experimentation from JAFFE database. The CK+ database consists of sequence of expressions in which each sequence starts with neutral and ends with a peak expression. In our experiments, we used the last image from each sequence where the expression is at its peak. Thus, 183 images from JAFFE and 328 images from CK+ were used in the experiment. As discussed earlier, the face from each image is aligned and brought to a common resolution of 96x96. The shape and appearance features from the facial patches are extracted. We have used five-fold cross validation method for evaluating the proposed technique. To accomplish this, we divided the database randomly into five parts containing 20% of dataset which was achieved by selecting 20% of data from each class randomly. Then, the OAO multi-class SVM classifier model is trained using any four parts of the data which is further used for testing of the images of the left out portion of the dataset. The average recognition rates for both the databases with five-fold cross validation are reported in Table I and Table II.

TABLE I. CONFUSION MATRIX OF CK+ DATABASE USING RBF SVM (AVERAGE RECOGNITION RATE = 94.63%)

|  | Anger | Fear | Disgust | Happiness | Sadness | Surprise |
|---|---|---|---|---|---|---|
| Anger | 95.26 | 0.53 | 1.05 | 1.58 | 1.58 | 0.00 |
| Fear | 1.25 | 88.75 | 0.00 | 10.00 | 0.00 | 0.00 |
| Disgust | 0.56 | 0.00 | 97.78 | 0.00 | 0.00 | 1.67 |
| Happiness | 0.00 | 2.50 | 0.00 | 95.50 | 0.00 | 2.00 |
| Sadness | 1.58 | 0.53 | 0.00 | 1.58 | 94.21 | 2.11 |
| Surprise | 0.53 | 0.00 | 0.53 | 1.58 | 1.05 | 96.32 |

TABLE II. CONFUSION MATRIX OF JAFFE DATABASE USING LINEAR SVM (AVERAGE RECOGNITION RATE = 87.43%)

|  | Anger | Fear | Disgust | Happiness | Sadness | Surprise |
|---|---|---|---|---|---|---|
| Anger | 98.33 | 1.67 | 0.00 | 0.00 | 0.00 | 0.00 |
| Fear | 3.33 | 88.33 | 5.00 | 0.00 | 1.67 | 1.67 |
| Disgust | 0.00 | 4.08 | 79.59 | 10.20 | 2.04 | 4.08 |
| Happiness | 0.00 | 1.67 | 3.33 | 93.33 | 1.67 | 0.00 |
| Sadness | 3.33 | 5.00 | 5.00 | 1.67 | 83.33 | 1.67 |
| Surprise | 0.00 | 1.67 | 5.00 | 5.00 | 6.67 | 81.67 |

TABLE III. AVERAGE RECOGNITION RATE IN BOTH DATABASES USING LINEAR, POLYNOMIAL, AND RBF SVM

| Database | Linear SVM | Polynomial SVM | RBF SVM |
|---|---|---|---|
| CK+ | 90.80 | 92.68 | **94.63** |
| JAFFE | **87.43** | 78.57 | 83.86 |

TABLE IV. COMPARISON OF RECOGNITION RATE IN BOTH DATABASES USING DIFFERENT FEATURES WITH RBF SVM

|  | CK+ | JAFFE |
|---|---|---|
| PHOG | 89.91% | 76.43% |
| LBP | 60.54% | 55.71% |
| PHOG+LBP | 94.63% | 83.86% |

As observed from Table I, the disgust is the easiest expression to recognize followed by surprise, happiness, and anger. In contrast, Table 2 shows that disgust is very difficult to be classified, whereas anger and happiness are easily classified. To conclude from Table I and II, happiness and anger has highest recognition rate in both the databases. We observed an overall accuracy of 87.43% for JAFFE database by using linear SVM. In CK+ database, it performed well with a recognition accuracy of 94.63%.

Table III depicts the accuracy when different kernels for SVM are used. We have conducted experiment on three kernels such as linear, polynomial, and RBF. The results show that RBF kernel performs best in CK+ database, whereas linear kernel performs well in JAFFE database. Table IV provides a comparison of performance of the individual features when used alone and the performance of the combined features. Here the individual feature vectors are obtained by concatenating the same features extracted from the defined facial patches. It was observed that the use of both shape and appearance features improve the accuracy of expression recognition.

## IV. CONCLUSION

In this paper, we have proposed an automated facial expression recognition method which uses combined shape and appearance features to encode the facial expressions. An active patch localization method is adopted by which unnecessary processing of the whole face region is avoided for expression recognition. This improves the performance while reducing computational cost. We used linear discriminant analysis for dimensionality reduction while preserving discriminative information of different expression classes. The performances of different SVM kernels are compared and it was found that both RBF and linear kernels are equally important depending on the nature of facial features. The proposed method performed well for both CK+ and JAFFE databases and promising accuracy was obtained in recognizing all expression classes of multiple subjects. Concatenation of shape and appearance features outperform the individual features in case of expression recognition. Though the methods are not evaluated for out-of-plane head rotation due to unavailability of such databases, we strongly believe that it will perform equally well. In future, we would like to evaluate the performance of such a system in real time applications.

ACKNOWLEDGMENT

The authors would like to thank Prof. Jeffery Cohn for providing the Cohn–Kanade database, and Dr. Michael J. Lyons for providing JAFFE database.